\documentclass[final]{cvpr}

\usepackage{times}
\usepackage{epsfig}
\usepackage{graphicx}
\usepackage{amsmath}
\usepackage{amssymb}
\usepackage{subcaption}
\usepackage{multirow}
\usepackage{algorithm}
\usepackage[noend]{algpseudocode}

\usepackage{nopageno}

\usepackage[pagebackref=true,breaklinks=true,colorlinks,bookmarks=false]{hyperref}

\def\cvprPaperID{10} 
\def\confYear{CVPR 2021}

\begin{document}


\title{Detecting and Matching Related Objects with One Proposal Multiple Predictions}

\author{Yang Liu \qquad Luiz G. Hafemann \qquad Michael Jamieson \qquad Mehrsan Javan\\
Sportlogiq, Montreal, Canada\\
{\tt\small \{yang, luiz, mikej, mehrsan\}@sportlogiq.com}
}

\maketitle

\begin{abstract}
Tracking players in sports videos is commonly done in a \emph{tracking-by-detection} framework, first detecting players in each frame, and then performing association over time. While for some sports tracking players is sufficient for game analysis, sports like hockey, tennis and polo may require additional detections, that include the object the player is holding (e.g. racket, stick). The baseline solution for this problem involves detecting these objects as separate classes, and matching them to player detections based on the intersection over union (IoU). This approach, however, leads to poor matching performance in crowded situations, as it does not model the relationship between players and objects. In this paper, we propose a simple yet efficient way to detect and match players and related objects at once without extra cost, by considering an implicit association for prediction of multiple objects through the same proposal box. We evaluate the method on a dataset of broadcast ice hockey videos, and also a new public dataset we introduce called COCO +Torso. On the ice hockey dataset, the proposed method boosts matching performance from 57.1\% to 81.4\%, while also improving the meanAP of player+stick detections from 68.4\% to 88.3\%. On the COCO +Torso dataset, we see matching improving from 47.9\% to 65.2\%. The COCO +Torso dataset, code and pre-trained models will be released at \url{https://github.com/foreverYoungGitHub/detect-and-match-related-objects}.

\end{abstract}

\section{Introduction}

Detecting players is the foundation of automated player tracking in sport games \cite{thomas_computer_2017,  komorowski_footandball_2020,  cioppa2020multimodal}, in a tracking-by-detection framework \cite{bewley_simple_2016, wojke_simple_2017, wang_towards_2020}. The detection task normally focuses on finding bounding boxes around players and, for some applications, a box around the ball. For sports like soccer or basketball, these detections can be sufficient for game analysis and estimation of players’ physical metrics \cite{manafifard_survey_2017}. In sports like hockey, polo and tennis, where players use a tool (e.g. a racket or a hockey stick), these detections are not sufficient for all applications. For these sports, it is also important to detect additional bounding boxes, either around the tools, or a larger bounding box around the player with the tool, and make the association to the player wielding it.

As an example, we consider a tracking system for hockey. For the application of classifying the type of a shot, the hockey stick needs to be tracked, and associated with the player holding it. In order to estimate the player's pose and stick location, a bounding box containing both the player and the stick can be used. On the other hand, for player identification (recognizing jersey numbers, or extracting Re-ID features), a bounding box only detecting the player is more appropriate. During crowded scenes (e.g. players fighting over a puck), players are close to each other, and the extended bounding box can contain part of another player, degrading performance for player identification.

The naive solution for this problem is to detect these boxes independently. This requires associating the regular and extended bounding boxes for the player, for instance using the Hungarian method. In crowded scenes, heuristics such maximizing intersection-over-union (IoU) or minimizing visual feature distance are not reliable for matching. Therefore, it is appealing to solve the problem of detecting the related boxes and matching them together at the same time in an end-to-end framework.

In this work, we propose a robust yet efficient method to extract both types of bounding boxes at the same time, where the association between them is implicit. This implicit association  comes at no extra cost, making it appealing for real-time tracking systems. 
Furthermore, the model can be trained without annotating the extended boxes in every image, thereby reducing the overall annotation effort.

The main contributions for this paper are:
\begin{itemize}
  \item We propose a method to simultaneously detect and match related bounding boxes. The related objects are naturally grouped by implicit association without extra cost. This includes the introduction of SetNMS to suppress duplicate sets of detections and to keep more possible detection sets in crowded scenes.
  \item We validate our proposed method on a broadcast ice hockey dataset and a public dataset we introduce called COCO +Torso dataset. Compared with the baseline model, the proposed methods performs better in both detection and matching for these datasets. Based on the ablation study, the proposed method is particularly suited for applications with highly overlapped associations.
\end{itemize}

\section{Related work}

\subsection{Object Detection in Sports}
Object detection has been widely used in automated tracking systems for sports \cite{thomas_computer_2017,  komorowski_footandball_2020,  cioppa2020multimodal, voeikov2020ttnet}. The task poses additional challenges to generic object detection, including camera distortion and fast movement (for broadcast videos), as well as frequent crowded scenes. Thaler \emph{et al.} \cite{marcus2013real} and Faulkner \emph{et al.} \cite{hayden2015AFL} adopt HOG and Haar features to detect and classify players in a sliding window for soccer and football. Acuna \cite{acuna2017towards} introduces an end-to-end CNN-based object detector in the basketball. In order to improve the detection result for fisheye cameras in soccer, Cioppa \emph{et al.} \cite{cioppa2020multimodal} applies a single stage object detector on cropped and rotated image patches. 



\subsection{Object Detection with Multiple Predictions}
Object detection with multiple predictions is a new approach for end-to-end CNN-based object detectors. The methods proposed in the literature are applied to two stage object detectors, with the goal of improving detector performance in crowded scenes. Chu \emph{et al.} \cite{chu2020detection} introduces multiple instance detection head to improve the object detection in the extremely high overlap scenario by predicting multiple unmatched objects from single proposal box. Zhang \emph{et al.} \cite{zhang2019double} and Huang \emph{et al.} \cite{huang2020nms} propose paired-RPN to predict the matched human visible part and human body with two proposal boxes to improve the pedestrian detection in crowd scenes. It is worth noting  that these methods were designed to improve human detection in the crowded scenes, rather than jointly solving the detection and matching problems. 

\begin{figure}[t]
\vspace{-0.3cm}
\includegraphics[width=.48\textwidth, trim={0.2in 0 0 0},clip]{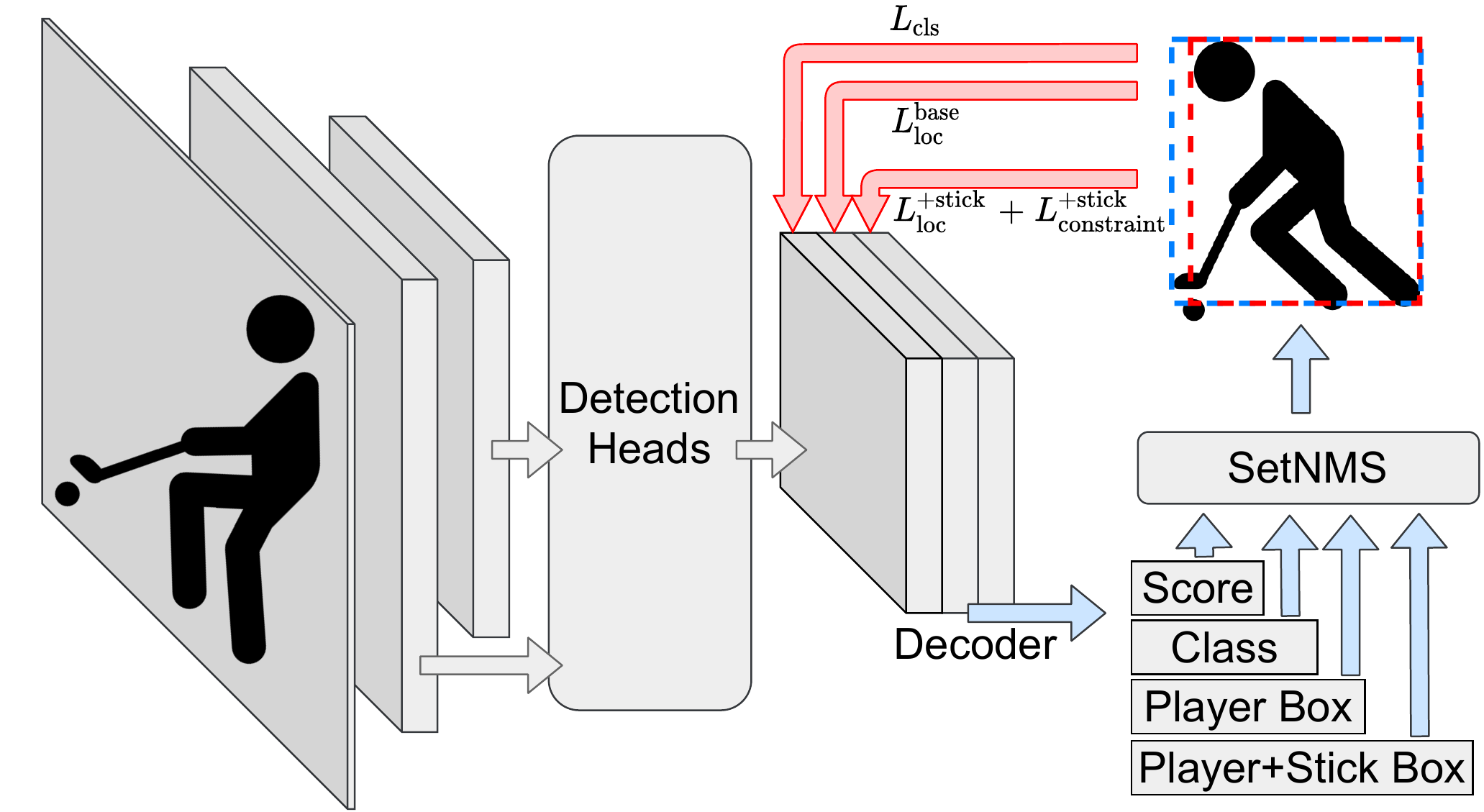}
   \caption{Illustration of the proposed network architecture for one proposal multiple prediction, applied to the ice hockey application. The arrows represent the data flow in the network: the red arrows indicate the loss functions applied to the networks during training, while the blue arrows shows the post-processing operations in inference.}
   \vspace{-0.3cm}
\label{fig:NetworkArchitecture}
\end{figure}

\section{Framework}
In order to detect and match related objects, we design the detector to output multiple predictions from a single proposal box, as shown in Figure \ref{fig:NetworkArchitecture}. 

During training, we introduce a specific loss function to jointly optimize the prediction of multiple associated objects from a single prediction box. 
For inference, we introduce the SetNMS procedure to suppress duplicated sets of detections. 


\subsection{Detection Head Design for One Proposal Multiple Predictions}

The loss functions in modern end-to-end detectors are based on proposal boxes: given a set of ground truth bounding boxes for an image, these are first matched to proposed boxes and then each proposal box is refined to its box location and target label. Besides not modeling relationships between boxes that belong to the same group, this creates another problem: each proposed box is forced to correspond to a single ground truth box, which creates an issue when multiple ground truth boxes are too close to each other. This is a common situation in the problems at hand. For example, the box for a player and for the player holding the stick normally have high overlap. 

We formulate the detection problem to jointly predict a set of related objects, such that the association between the objects is implicit, addressing both the issue of associating bounding boxes, and the problem of training a detector when multiple ground-truth boxes overlap. We consider $b_\text{base} = \{x_\text{base}, y_\text{base}, w_\text{base}, h_\text{base}\}$ the \emph{base} bounding box (e.g. for the player), and a sequence of $N$ associated bounding boxes $B_\text{extra} = \{b_i | 1 \le i \le N \}$. We model the associated bounding boxes as \emph{offsets} from the base box. That is, for the extra box $i$, the model estimates $\{x^i_\text{offset}, y^i_\text{offset}, w^i_\text{offset}, h^i_\text{offset}\}$, and the bounding box is specified as $b_i = \{x_\text{base} + x^i_\text{offset}, y_\text{base} + y^i_\text{offset}, w_\text{base} + w^i_\text{offset}, h_\text{base} + h^i_\text{offset} \}$. Finally, given an input image $X$, the detector estimates a set of paired (matched) boxes $\{b_\text{base}\} \cup B_\text{extra}$, together with a class $c$ (for the whole set) and and an objectness score $s$.

This formulation assumes that the related objects can be predicted from the base object. Based on this assumption, the proposal boxes are first matched to the objects in the base class, and the related objects are matched to the same proposal boxes. Therefore, the same proposal box is used for the  group of related objects. Since the predicted objects are already matched, there is only one predicted score and class for each matched group. Additionally, the problem for high overlapped ground truth bounding boxes (for objects on the same group) is solved, as the related objects share the same proposal box, instead of competing for it. It should be noticed that, even though the detection head we propose is designed on top of single-stage detectors like Yolo/FPN, it can be easily extended to other two stage detector frameworks like Faster R-CNN and Mask R-CNN.




\subsection{Loss Functions}


We consider the following loss function:

\begin{equation}
\label{eq:loss}
L_\text{total} = L_\text{cls} + L_\text{loc}^\text{base} + \sum_{i=1}^{n}(L_\text{loc}^{i}+L_\text{constraint}^{i})   
\end{equation}

Where $L_\text{cls}$ is the classification loss for the set. In this work we use the focal loss \cite{lin_focal_2017};
$L_\text{loc}^\text{base}$ and $L_\text{loc}^{i}$ are the localization losses for the base box and extra boxes, respectively, that are defined below. Finally, $L_\text{constraint}^{i}$ is a term to represent the relationship between boxes from the base classes and the related classes $i$. This is an optional term, that we can use to model domain knowledge for a particular application, as described below and in \autoref{eq:constraint}. 

For the localization losses, we use the Distance-IoU Loss \cite{zheng2020distance},  defined as follows:


\begin{equation}
\label{eq:diou}
    L_\text{loc} = 1 - \text{IoU} + \frac{||c - c_\text{gt}||^2}{d^2}
\end{equation}
Where IoU indicate the intersection-over-union of the predicted and ground-truth boxes,  $c$ and $c_\text{gt}$ are the center points of the predicted and GT boxes (respectively), $d$ is the diagonal length of the smallest enclosing box that covers the two boxes.

We also model an optional constraint loss, that can be used to model relationships between base and extra bounding boxes. In particular, we consider a term to encourage that an extra bounding box $b_i$ be strictly larger than the base bounding $b_\text{base}$. This term is appropriate for situations where the extra bounding boxes are intended to always enclose the base bounding box. This loss term is defined by adding a penalty if the extra bounding box does not cover the base bounding box. We use the following loss for the $x$ component:

\begin{equation}
\label{eq:constraint}
L^i_\text{constraint\_x} = 
\begin{cases}
  \text{smooth}_{L1}(x_\text{base} - x_i), & \text{if}\ x_i > x_\text{base} \\
  0, & \text{otherwise}
\end{cases}
\end{equation}
Where smooth$_{L1}$ is the Smooth L1 loss defined in \cite{girshick_fast_2015}. We use equivalent losses for $y$, $x + w$ and $y + h$, penalizing the following conditions: $y_i > y_\text{base}$, $x_i + w_i < x_\text{base} + w_\text{base}$,  and $y_i + h_i < y_\text{base} + h_\text{base}$.


\begin{algorithm}[]
 \hspace*{\algorithmicindent} \textbf{Input} \\
 \hspace*{\algorithmicindent}    $B = \left \{ \left \{b_1^{1},...,b_1^{C}\right \},...,\left \{b_N^{1},...,b_N^{C}\right \}\right \}$: boxes. \\
 \hspace*{\algorithmicindent}    $S = \left \{s_1,...,s_N\right \}$: scores. \\
 \hspace*{\algorithmicindent}    $C = \left \{c_1,...,c_N\right \}$: classes. \\
 \hspace*{\algorithmicindent}    $\Omega_{IoU}$: IoU threshold for SetNMS.
 \begin{algorithmic}[1]
 \Procedure{SetNMS}{$B,S,C,\Omega_{IoU}$}
    \State $R \gets \left \{ \right \}$
    \While {$B \neq \emptyset$}
        \State Let $T$ be the highest scored group 
        \State Remove $T$ from $B$, $S$ and $C$, add it to $R$
        \For{$\left \{b_i^{1},...,b_i^{C}\right \} \in B$}
            \If {$c_i \neq c_T$}
                Continue
            \EndIf
            \State $IoU_{set} \gets \left \{ IoU(b_T^{1},b_i^{1}),...,IoU(b_T^{C},b_i^{C}) \right \}$
            \If {$\min (IoU_{set}) > \Omega_{IoU}$}
                \State Remove $i$-th element from $B$, $S$, $C$
            \EndIf
        \EndFor
    \EndWhile
    \Return $R$
 \EndProcedure
 \end{algorithmic}
 \caption{SetNMS}
 \label{alg:setnms}
\end{algorithm}

\begin{figure*}[]
\centering
\includegraphics[width=.95\textwidth, trim={0 0 0 0},clip]{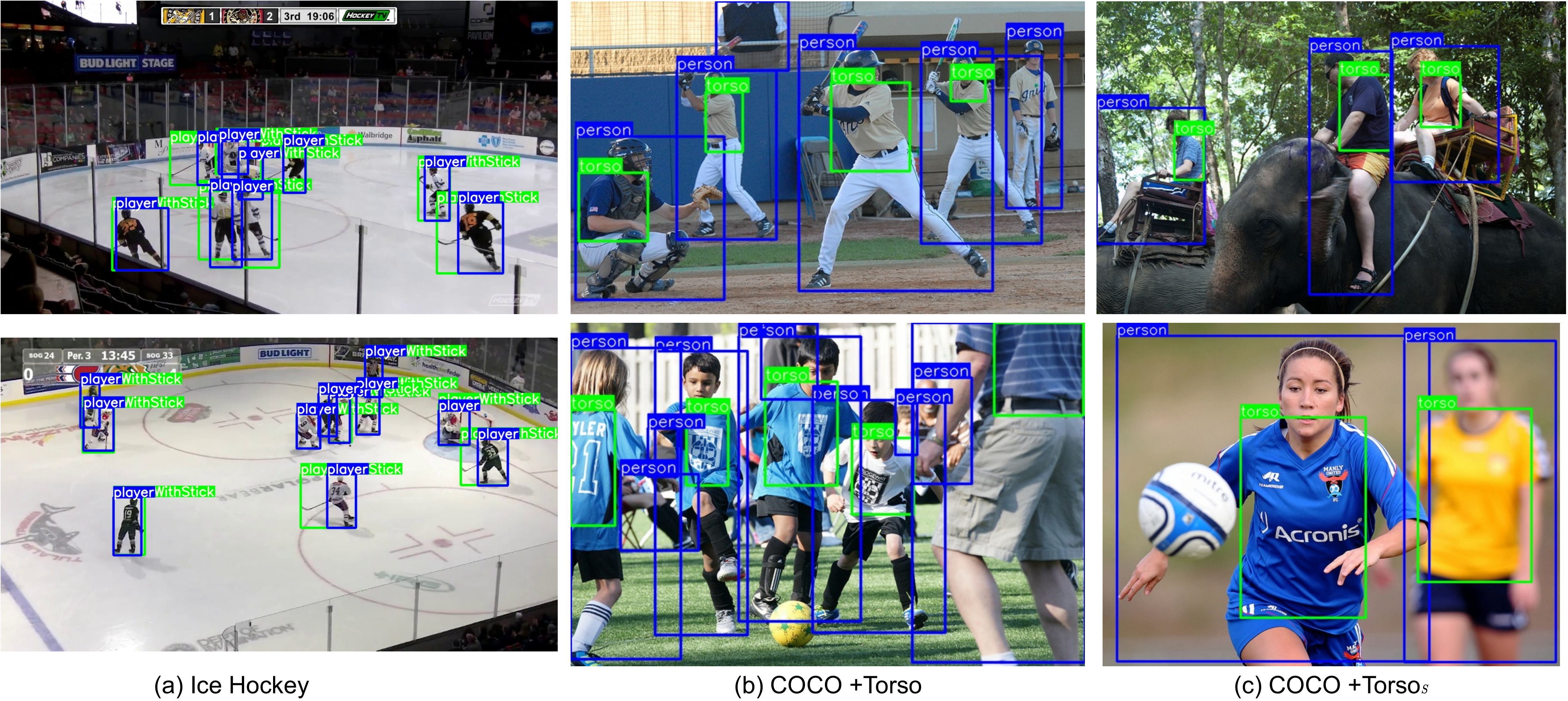}
   \caption{Samples from the Ice hockey, COCO +Torso and COCO +Torso$_S$ datasets. The blue boxes indicate the detections for the base class, which is the player class in ice hockey dataset and person class in COCO +Torso dataset, while the green boxes are the bounding boxes for the related classes, which is the player+stick class in ice hockey dataset and torso class in COCO +Torso dataset.}
\label{fig:datasets}
\end{figure*}

\subsection{SetNMS}

During inference time, detectors normally use a non-maximum suppression (NMS) algorithm to avoid duplicated predictions to the same object. In the method above, we predict a set of matched boxes for each proposal, and therefore we need to adapt this procedure to work with a set of predictions - we name this process SetNMS.


The SetNMS algorithm is described in Algorithm \ref{alg:setnms}. Since the predicted matched group shares the same classification result and confidence score, SetNMS suppresses false positive predictions by taking all the matched boxes in the group into consideration rather than taking one predicted box. Only all the matched boxes in the group that have high overlap with the boxes in the kept group will be considered as false positive predictions and be removed. Since the SetNMS compares the ensemble group of the matched boxes rather than a single box, it also keeps more reliable boxes in the crowd scene and more robust for the whole ensemble group compared to the original NMS.

\section{Experiments}
In this section, we discuss our evaluation protocol, and discuss the results of our approach on our ice hockey dataset and the COCO +Torso dataset.

\subsection{Datasets and Evaluation Metric}
\textbf{Ice hockey dataset} is a dataset for ice hockey player detection in broadcast videos. The images are sampled from multiple ice hockey leagues including the National Hockey League (NHL) and minor leagues. Figure \ref{fig:datasets} shows two images of this dataset. Two classes are considered: “player” - a tight bounding box around the player, and “player+stick” - a bounding box that includes the player and the hockey stick. Only part of the images contains the “player+stick” label. This dataset contains 15095 images in the training set and 1886 images in the validation set. Of these, only 3286 images in the training set and 1634 images in the validation set have the annotation for the “player+stick” class. For the samples where the player+stick annotation is not available, we simply remove the localization loss for the related classes in Equation \ref{eq:loss}. Note that we always consider the constraint loss, that drives the model to predict player+stick bounding boxes that are always larger than the player bounding boxes, even when we do not have annotations for player+stick.  We benchmark the proposed method to predict both classes. 

\textbf{COCO +Torso dataset} is a public dataset introduced in this work, based on the COCO\cite{lin2014microsoft} and DensePose\cite{guler2018densepose} datasets. The DensePose dataset is mainly used to predict dense human body joints and contains 3 labels: human bounding boxes, human body joints and human densepose. Based on these three labels, we generated the COCO +Torso dataset featuring detections for two classes: a ``person'' class with their matched ``torso''. The human body joints are used to filter the human bounding boxes which does not contains the any torso joints annotation, while the DensePose annotation is used to generate the torso bounding boxes. Two sample images in this dataset are shown in Figure \ref{fig:datasets} (b). The training and validation sets for COCO +Torso dataset are split based on its original split set in the DensePose dataset. The final COCO +Torso dataset contains 26437 and 5984 images in training and validation set respectively. 

\textbf{COCO +Torso$_S$ dataset} In order to simplify the COCO person-torso detection and matching task, we also sample an easier dataset from COCO +Torso dataset and named it as COCO +Torso$_S$ dataset. This simplified dataset only contains the images which all the persons in the image have the matched torso bounding boxes. Two sample images in this dataset is shown in Figure \ref{fig:datasets} (c). After sampling, the training and validation sets for the COCO +Torso$_S$ dataset contain 13483 and 2215 images separately. 


\begin{table*}[]
\centering
\begin{tabular}{cc|ccc|ccc}
\hline
Dataset                           & Method           & \multicolumn{3}{c|}{$AP^{0.5} (\uparrow)$} & \multicolumn{3}{c}{$MR^{-2} (\downarrow)$}  \\ \hline
                                  &                  & Player        & Player+Stick  & Match         & Player       & Player+Stick  & Match \\ \hline
\multirow{3}{*}{Ice Hockey}       & Yolo Player Only & 98.9          &               &               & 1.6          &               &       \\
                                  & Yolo             & 98.5          & 68.4          & 57.1          & 1.9          & 52            & 63.1    \\
                                  & Yolo+MP          & \textbf{99}   & \textbf{88.3} & \textbf{81.4} & \textbf{1.5} & \textbf{20.8} & \textbf{33.2}  \\ \hline
                                  &                  & Person        & Torso         & Match         & Person       & Torso         & Match \\ \hline
\multirow{4}{*}{COCO +Torso}      & Yolo             & 72.7          & 67.6          & 47.9          & 34.2         & \textbf{45.6} & 71.8  \\
                                  & Yolo+MP          & \textbf{74.7} & 66.1          & \textbf{65.2} & \textbf{29.2}& 48.5          & \textbf{52.1}  \\
                                  & FPN              & 71            & \textbf{67.7} & 45.9          & 35.7         & 46.1          & 73.3  \\
                                  & FPN+MP           & 72.7          & 64.4          & 63.3          & 30.7         & 48.9          & 52.3  \\ \hline
\multirow{4}{*}{COCO +Torso$_S$}  & Yolo             & 97.1          & 90.5          & 69.3          & 4.6          & 15            & 51.1  \\
                                  & Yolo+MP          & \textbf{98.1} & 87.3          & 86.1          & \textbf{2.7} & 21.5          & 24.1  \\
                                  & FPN              & 96.3          & \textbf{91.2} & 68.3          & 5.5          & \textbf{13.8} & 51.2  \\
                                  & FPN+MP           & 97.7          & 87.7          & \textbf{87.1} & 3.1          & 21.5          & \textbf{22.8}  \\ \hline
\end{tabular}
\caption{Detection and matching results on the three datasets. \textbf{+MP} indicates using the detection head and postprocessing methods proposed in this paper.}
\label{tab:result}
\end{table*}

\textbf{Evaluation Metric}. Since we consider a joint detection and matching task, the standard log-average miss rate ($MR$) \cite{dollar2011pedestrian} and average precision ($AP^{.50}$) \cite{everingham2015pascal} are adopted as detection evaluation metrics for each class. The $MR$ is computed in the false positive per image (FPPI) with a range of $[10^{-2}, 100] (MR^{-2} )$, while the $AP^{0.5}$ calculates the average precision with IoU threshold 0.5 with a range of $[0, 1]$. In order to evaluate model performance in object matching, we also extend the log-average miss rate and average precision and introduce the matching log-average miss rate (${MR}_{match}$) and average precision ($AP^{.50}_{match}$). Compared with the standard $MR$ and $AP^{.50}$, in ${MR}_{match}$ and $AP^{.50}_{match}$, the true positive ($TP$) is counted only if the IoU score for all the objects in the group is higher than IoU threshold. The range of $AP^{.50}_{match}$ is in $[0, \min (AP^{.50}_{1},...,AP^{.50}_{C})]$, while the range of ${MR}_{match}$ is in $[\max ({MR}_{1},...,{MR}_{C}), 100]$, where $C$ is the classes for the grouped objects.

\subsection{Experimental protocol}
In these experiments, we adopt the state-of-the-art single stage object detector pretrained at COCO dataset \cite{lin2014microsoft} as our baseline.
The feature extractor in the baseline models is ResNet-18\cite{he2016deep}, while YOLOv3\cite{redmon2018yolov3} and FPN\cite{lin_focal_2017} are used as the detection head in the baseline models. It should be noted that there is only 1 anchor box for each proposal region in the anchor-based single stage detector. The model input size for ice hockey dataset is 1280x720, while the input size for COCO +Torso datasets is 320x320.

The same training strategy is applied for all experiments: we first train only the detection head for 60 epochs, using Adam with learning rate $4 \times 10^{-4}$ and then fine-tune the whole network for 250 epochs using Adam with learning rate $10^{-4}$. During the training, we adopt center sampling and scale matching to match the proposal boxes with ground truth bounding box. We also use standard data augmentation methods: random resizing, cropping, flipping, random contrast and saturation. We do not employ multi-scale predictions during training or testing.

\subsection{Results}

\textbf{Ice hockey dataset}: We consider two baselines for this dataset: a detector trained only for the player class and a detector trained on the full dataset, but treating the two classes as separate. For the latter, we match the player and player+stick predictions at inference time using the Hungarian algorithm, based on the IoU and confidence score between boxes of the two classes. We compare these baselines to the proposed method that outputs detections for both classes from the same proposal, which performs the matching implicitly. The results are shown in Table \ref{tab:result}. We can see that the baseline detector achieves 98.9 AP in player class when it is only trained for this class. When the extra player+stick class is introduced to the baseline detector, the $AP_{player}$ drops 0.4 and the $AP_{player+stick}$ only achieves 68.4. We hypothesize that this worse performance is caused by the the two classes competing for the same proposal boxes in situations with high overlap between the two classes. The proposed detector solves this issue by design, and achieve similar accuracy for the Player class, while significantly boosting performance for the Player+stick class to 88.3\%.

We notice a more significant difference on the the matching performance ($AP_{match}$ and $MR_{match}$). The $AP_{match}$ for the baseline method is 57.1 which is 83.4\% of its upper bound ($AP$ for the player+stick class), while the $AP_{match}$ for the proposed detector achieves 81.4\%, which is the 92.2\% of its upper bound. The $MR_{match}$ for the proposed detector is also 1.90 times lower than the baseline detector as well, showing that the proposed detector the best result in all detection and matching evaluation metrics.


\textbf{COCO +Torso dataset}: In order to check the generalization of our proposed method on non-sport tasks, we trained and evaluated the proposed framework in the COCO +Torso dataset to detect the bounding boxes for the person and torso classes and matched the corresponding boxes for each person. The result for this experiment is shown in Table \ref{tab:result}. Compared with the baseline models, the proposed detector with one proposal multiple prediction head improves $AP_{person}$ by 1-2 percent points, but decreases $AP_{torso}$ by 1.5-3.5 percent points. We notice, however, a significant improvement in matching performance: $AP_{match}$ has 1.36 and 1.24 improvement for COCO +Torso and COCO +Torso$_S$ dataset respectively. This result shows that the proposed framework is suitable for applications where the \emph{matching} task is important.


\begin{figure}[t]
\vspace{-0.2cm}
\centering
\includegraphics[width=.475\textwidth, trim={0.11in 0.3in 0.1in 0.1in}, clip]{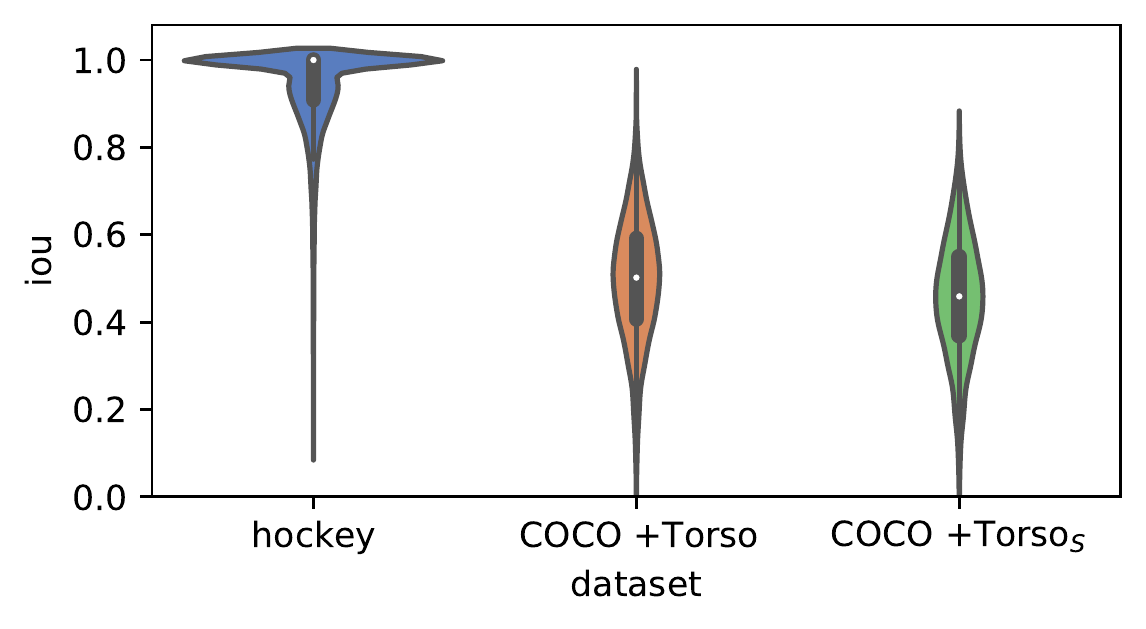}
  \caption{Density distribution for IoU scores between the base objects and related objects among ice hockey, COCO +Torso and COCO +Torso$_S$ dataset.}
  \vspace{-0.3cm}
\label{fig:iou_density_distribution}
\end{figure}

\begin{figure}[b]
\vspace{-0.3cm}
\centering
     \begin{subfigure}[b]{0.15\textwidth}
         \centering
         \includegraphics[width=\textwidth, trim={0.1in 0 0 0},clip]{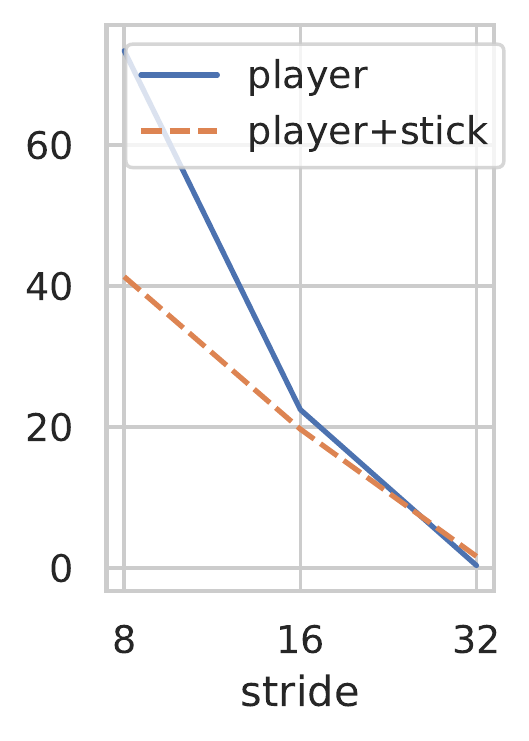}
         \caption{Ice Hockey}
         \label{fig:match_ice}
     \end{subfigure}
     \hfill
     \begin{subfigure}[b]{0.15\textwidth}
         \centering
         \includegraphics[width=\textwidth]{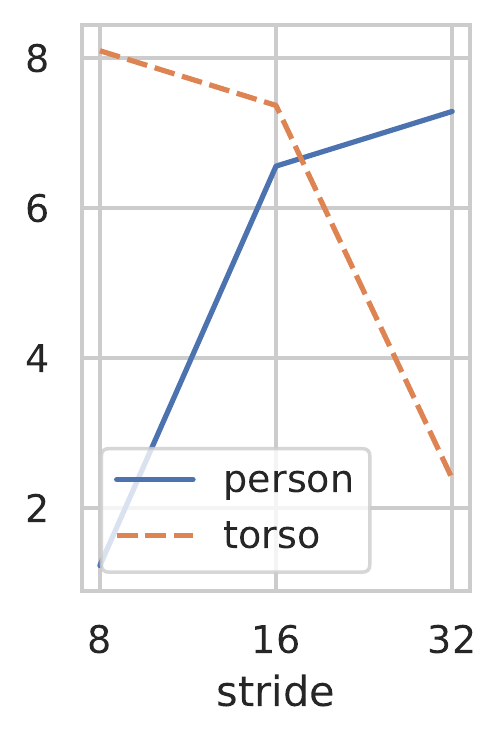}
         \caption{COCO +Torso}
         \label{fig:match_easy}
     \end{subfigure}
     \hfill
     \begin{subfigure}[b]{0.15\textwidth}
         \centering
         \includegraphics[width=\textwidth]{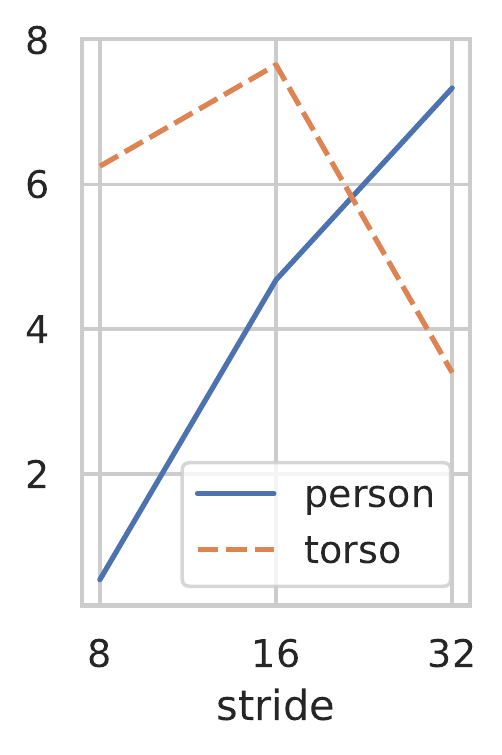}
         \caption{COCO +Torso$_S$}
         \label{fig:match_hard}
     \end{subfigure}
\caption{The number of proposal boxes matched per image in each feature map for the three datasets. The $x$ axis is the stride between the feature map and original input size, while the $y$ axis is the number of matched proposal boxes for each class.}
\label{fig:match_three}
\vspace{-0.3cm}
\end{figure}

\textbf{Ablation Study on Proposal Boxes Matching}: The results from Table \ref{tab:result} show that the proposed method achieves better results on base class detection and group matching in all experiments. However, while the proposed model trained with ice hockey dataset has significant improvement in the related classes, the same model trained on the COCO +Torso dataset performs worse than the regular detector for the extra ``Torso'' class. In order to obtain insights on which situation the proposed framework works better, we conducted two ablation studies: we first analyze the distribution of IoU scores between base and extra bounding boxes (Figure \ref{fig:iou_density_distribution}). In a second experiment, we compute the average number of proposal boxes matches per image, among all feature maps (Figure \ref{fig:match_three}). 

Figure \ref{fig:iou_density_distribution} shows the  distribution of the IoU scores for matched objects in the three datasets (e.g. the IoU between a player and the associated box of player+stick). We notice a very different distribution for the ice hockey dataset, where the overlap between the two classes is very high (mean of 0.94), whereas for the COCO +Torso datasets, we see a mean around 0.5 (0.46 for COCO +Torso$_S$).
Since the IoU score of paired objects is only 0.5 in the COCO +Torso dataset, in the regular detector, the groundtruth boxes can be assigned with the suitable proposal boxes based on its center and size. However, for ice hockey dataset, since the matched objects are highly overlapped with each other, these objects almost always need to compete for the same proposal boxes in regular detector. 
Therefore, the proposed method is particularly suited for tasks with high-overlap between the base bounding box and extra boxes.

Figure \ref{fig:match_three} shows the number of proposal box matches, per feature map, in the three different resolutions considered by the detection heads (for stride $s$ the feature map has size $\frac{h}{s} \times \frac{w}{s}$). The left plot (a) shows that the player and player+stick tend to match to the proposal boxes located in the same feature maps in ice hockey dataset. This indicates that the same feature maps should contain the rich information for both matched classes in this dataset. However, in COCO +Torso dataset, since the torso objects are much smaller than the person objects, the paired objects are matched to the proposal boxes in the different feature maps in the regular detectors (plots (b) and (c)). In the proposed method, the proposal boxes are matched to the bounding box of the \emph{base} objects, and its feature map may not be suitable for the related matched objects. Therefore, the proposed detector performs slightly worse in the torso classes in COCO +Torso dataset. This reinforces the previous finding that the proposed method is particularly suited for applications where the base and extra classes have high overlap.

\begin{table}[t]
\centering
\begin{tabular}{cc|ccc}
\hline
Dataset                           & Method           & \multicolumn{3}{c}{$AP^{0.5} (\uparrow)$} \\ \hline
                                  &                  & Player  & +Stick  & Match \\ \hline
\multirow{3}{*}{Ice Hockey}       & NMS              & 98.9    & 88.1               & 81.2      \\
                                  & JointNMS\cite{zhang2019double} & 98.8    & 88          & 81  \\
                                  & SetNMS           & \textbf{99}      & \textbf{88.3}          & \textbf{81.4}  \\ \hline
                                  &                  & Person  & Torso         & Match \\ \hline
\multirow{4}{0.9cm}{\centering COCO +Torso}           & NMS              & \textbf{76.4}    & \textbf{66.7}          & \textbf{66.5}     \\
                                  & JointNMS\cite{zhang2019double}  & 76.3    & 66.6          & 65.3  \\
                                  & SetNMS           & 74.7    & 66.1          & 65.2  \\ \hline
\multirow{4}{0.9cm}{\centering COCO +Torso$_S$}       & NMS              & \textbf{98.4}    & 87            & 85.7  \\
                                  & JointNMS\cite{zhang2019double}  & 98.2    & 86.9          & 85.4  \\
                                  & SetNMS           & 98.1    & \textbf{87.3}          & \textbf{86.1}  \\ \hline
\end{tabular}
\caption{Ablation studies on NMS methods for the one proposal multiple prediction model.}
\label{tab:nms}
\vspace{-0.3cm}
\end{table}

\textbf{Ablation Study on NMS methods}: In order to validate the proposed SetNMS, we conducted an ablation study with three different NMS methods: (i) the baseline (denoted simply by NMS) considers performing NMS only with the IoU score of the base class; (ii) JointNMS\cite{zhang2019double} removes group of objects as long as objects in \emph{one class} are overlapped; (iii) the proposed SetNMS, which removes the group only if all of objects in the group are overlapped with other groups. The result of this comparison is shown in Table \ref{tab:nms}.
 
Based on its formulation, the proposed SetNMS should keep the highly overlapped groups in a crowded scene, but also introduce more false alarm groups with low confidence scores. In Table \ref{tab:nms}, we see that the results for different methods differ only by less than 1 percent point. SetNMS achieves the best result in all detection and matching evaluation metrics for Ice Hockey dataset, while NMS achieves the best result for COCO +Torso dataset. For COCO +Torso$_S$ dataset, NMS has better AP in person detection while the best torso detection and objects match AP is generated by SetNMS. This result indicates that when the model is not confident about the predicting boxes, the SetNMS actually performs worse than the normal NMS. Therefore, the choice of the best NMS procedure can be application-specific. 

\section{Conclusion}

In this paper, we propose an efficient object detector that can simultaneously detect and match player bounding boxes and related bounding boxes (e.g. player holding a stick), that can be used for applications where detecting just the player is not sufficient. This method considers an implicit association for the multiple predictions through the same proposal box to detect and match the related objects without extra cost. Our experiments show that this method is particularly suited for applications that require high matching accuracy, and situations where the base and related bounding boxes have high overlap.

{\small
\bibliographystyle{ieee_fullname}
\bibliography{egbib}
}

\end{document}